\documentclass{research4cacm}
\usepackage{amsmath}
\usepackage{mathtools}
\usepackage{multirow}
\usepackage{xcolor}
\usepackage{makecell}
\usepackage{graphics}
\usepackage{adjustbox}
\usepackage{enumitem,booktabs,cfr-lm}
\usepackage[referable]{threeparttablex}
\usepackage{tabularx}
\usepackage{array}
\usepackage{amssymb}
\usepackage[ruled,linesnumbered]{algorithm2e}
\usepackage{url}
\usepackage{diagbox}

\renewlist{tablenotes}{enumerate}{1}
\makeatletter
\setlist[tablenotes]{label=\tnote{\alph*},ref=\alph*,itemsep=\z@,topsep=\z@skip,partopsep=\z@skip,parsep=\z@,itemindent=\z@,labelindent=\tabcolsep,labelsep=.2em,leftmargin=*,align=left,before={\footnotesize}}
\makeatother

\DeclareMathOperator*{\argmin}{\arg\min}
\DeclareMathOperator*{\argmax}{\arg\max}

\newcommand{\hyper}{hyperparameter }
\newcommand{\hypers}{hyperparameters }
\newcommand{\Hyper}{Hyperparameter}

\begin{document}
%

\title{Techniques for Automated Machine Learning
%
\thanks{}
}


%
%
%
%
%
\numberofauthors{3} 
%

\author{
Yi-Wei Chen, Qingquan Song, Xia Hu \\
Department of Computer Science and Engineering \\
Texas A\&M University\\
\{yiwei\_chen, song\_3134, xiahu\}@tamu.edu
}

\maketitle
\begin{abstract}
Automated machine learning (AutoML) aims to find optimal machine learning solutions automatically given a machine learning problem. 
It could release the burden of data scientists from the multifarious manual tuning process and enable the access of domain experts to the off-the-shelf machine learning solutions without extensive experience.
In this paper, we review the current developments of AutoML in terms of three categories, automated feature engineering (AutoFE), automated model and \hyper learning (AutoMHL), and automated deep learning (AutoDL).
State-of-the-art techniques adopted in the three categories are presented, including Bayesian optimization, reinforcement learning, evolutionary algorithm, and gradient-based approaches. We summarize popular AutoML frameworks and conclude with current open challenges of AutoML.
\end{abstract}

\section{Introduction}

Automated machine learning (AutoML) has emerged as a prevailing research field upon the ubiquitous adoption of machine learning techniques. 
It aims at automatically determining high-performance machine learning solutions with a little workforce in reasonable time budget.
For example,
Google HyperTune, Amazon Model Tuning, and Microsoft Azure AutoML 
all provide cloud services with AutoML tools which cultivate off-the-shelf machine learning solutions for both researchers and practitioners.
Therefore, AutoML not only liberates them from the time-consuming tuning process and tedious trial-and-error iterations but also facilitates the development of solving machine learning problems.
 
\begin{figure}
    \centering
    \includegraphics[width=0.49\textwidth]{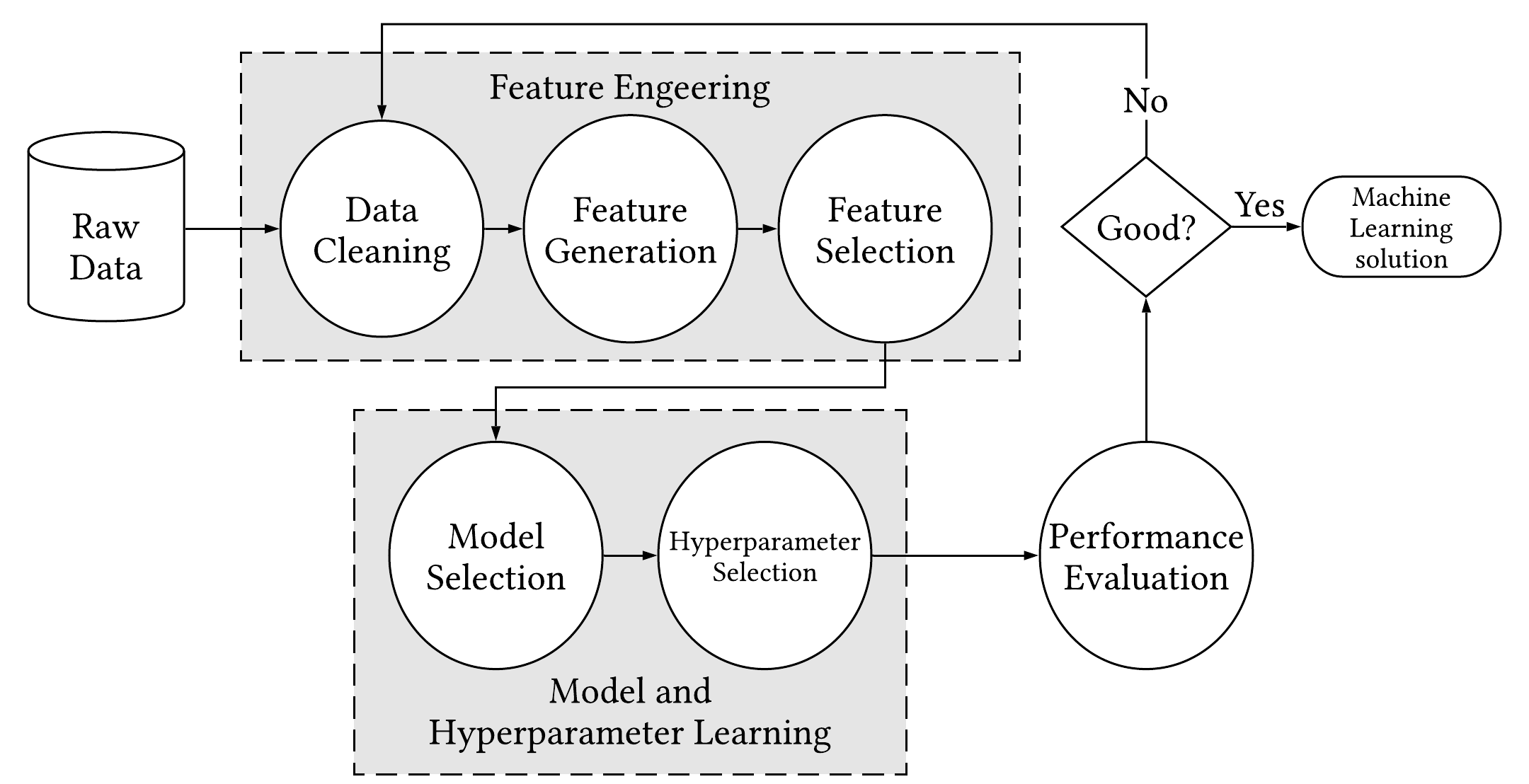}
    \caption{
    A classical machine learning pipeline. 
    When data scientists intend to find their machine learning solutions,
    they manually try the combination of features, models, and \hypers in a pipeline and repeatedly tune its performance until the final solution achieves good general prediction. 
    AutoML is proposed to facilitate the labor-intensive tuning process.
    }
    \label{fig:pipeline}
\end{figure}

\begin{figure}
    \centering
     \includegraphics[width=0.45\textwidth]{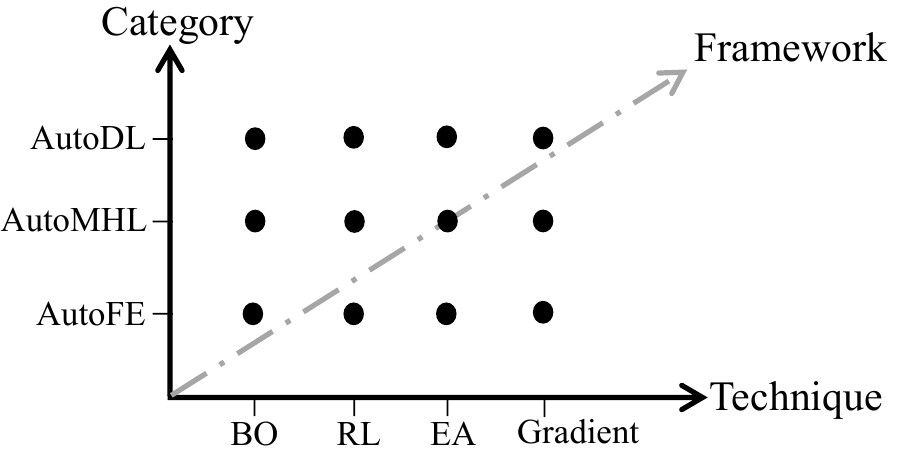}
    \caption{A taxonomy of AutoML from techniques and frameworks. 
        We split AutoML into automated feature engineering (AutoFE), 
        automated model and \hyper learning (AutoMHL), 
        and automated deep learning (AutoDL). 
        The three categories are illustrated from the techniques dimension, 
        including Bayesian optimization (BO), reinforcement learning (RL), evolutionary algorithm (EA), 
        and gradient-based approaches (Gradient). 
        Selected AutoML frameworks are discussed beyond.
    }
    \label{fig:taxonomy}
\end{figure}

A traditional machine learning pipeline is an iterative procedure composed of feature engineering, model and \hyper selection, and performance evaluation, as shown in Figure~\ref{fig:pipeline}. 
Data scientists manually manipulate numerous features, design models, and tune hyperparameters in order to get the desired predictive performance. 
The procedure will not be terminated until a satisfactory performance is achieved. 
Echoing with the traditional pipeline, we split AutoML into three categories, 
(1) AutoFE: automated feature engineering, 
(2) AutoMHL: automated model and \hyper learning, and 
(3) AutoDL: automated deep learning. 
AutoFE intends to discover informative and discriminative features for the learning model. 
AutoMHL aims at tuning \hyper of a specific learning model (HPO) or building an entire machine learning pipeline automatically (Auto-Pipeline). 
AutoDL is a subfield of AutoMHL. 
Since deep learning has succeeded significantly without intensive feature engineering,
we explicitly separate AutoDL from AutoMHL in order to focus on the automatic design of deep neural network architectures. 
Upon the categorization of AutoML, 
we will review AutoML from two dimensions (Figure~\ref{fig:taxonomy}). The technique dimension spans the mainstream AutoML techniques, including Bayesian optimization (BO), reinforcement learning (RL), evolutionary algorithm (EA), and gradient-based approaches (Gradient), while the framework dimension covers representative AutoML frameworks.


The remainder of the paper is organized as follows: 
Section~\ref{sec:autofe} discusses AutoFE with RL, EA, and other techniques. 
Section~\ref{sec: automhl} explains AutoMHL with BO, RL, and EA. 
Section~\ref{sec:autodl} illustrates AutoDL in terms of BO, RL, and Gradient. 
Section~\ref{sec:frameworks} explores the emblematic open-source and enterprise AutoML frameworks.
The last two sections specify current research challenges and reemphasize the target of this work.
Table~\ref{table:comparison_auto_ml} summarizes the AutoML techniques mentioned in the work.

\section{Automated Feature Engineering}
\label{sec:autofe}
Feature engineering is a process to manipulate features via operations such as data imputation, feature transformation, and feature selection.
It is a critical step in machine learning algorithms since suitable features directly influence their prediction performance~\cite{domingos2012few}. 
Considering a dataset $\mathcal{D}^F$ represented in terms of its feature set $F$,
the derived features $\hat{F}$ in $\mathcal{D}^{\hat{F}}$ are generated by applying feature pre-processing operations $\mathcal{T}$ to $\mathcal{D}^F$.
The automated feature engineering could be formulated in the following optimization problem, 
\begin{equation}
    \label{eq:atuofe}
    \begin{aligned}
        F^{*} & = \argmin_{F \in \mathcal{F} \cup \hat{\mathcal{F}}} \mathcal{L}_{val}(A_{w^*}, D_{val}^{F}),
        \\
        \text{s.t.}\ \  w^* & = \argmin_{w} \mathcal{L}_{train}(A_{w}, D_{train}^{F}),
    \end{aligned}
\end{equation}
where $\mathcal{L}_{train}$ and $\mathcal{L}_{val}$ are the loss function for training and validation,
$D_{train}^F$ and $D_{val}^F$ denote training and validation set from $D^F$, and 
$w$ is the parameters (weights) of a machine learning algorithm $A$.
We intend to pick up vigorous features that can boost the performance of $A$ through AutoFE. 
Generally, AutoFE follows the iterative procedure to find the optimal features, including two necessary phases, (1) feature generation and (2) feature selection.
The original features are extended by feature generation and then feature selection will filter irrelevant and overlapping features. 
In this section, we briefly introduce two popular techniques in AutoML, i.e., reinforcement learning (RL) and evolutionary algorithm (EA), under the context of AutoFE.

\begin{algorithm}[]
    \label{algo:rl}
    \SetAlgoLined
    \SetKwInOut{Input}{Input}
    \SetKwInOut{Output}{Output}
    \Input{The policy $\pi$; environment $env$, including the reward function;}
    \Output{The well-trained policy $\pi$}
     Initialize state $s_i$ and policy $\pi$ \\
     \While{not meet stopping criteria}{
        action $a^* = \argmax_a\pi(s_i, a)$ \\
        $s_{i+1}, r = env(s_i, a^*)$ \\
        update $\pi$ with $s_{i+1}, r$ \\
        i = i + 1 \\
     }
     \Return $\pi$
     \caption{Reinforcement Learning}
\end{algorithm}

\subsection{Reinforcement Learning}
\label{subsec:autofe_rl}
Reinforcement learning (RL) is one of the most popular techniques used in AutoML. The goal of RL is to train an agent that could learn how to take a sequence of proper actions to maximize the long-term reward in the specific environment. 
Algorithm~\ref{algo:rl} illustrates the iterative procedure of RL. 
The agent takes actions upon a particular strategy (a.k.a. policy) to learn experience for correct decisions. 
The environment produces the reward based on the action and the current state of the agent. 
The agent, the environment, actions, and the reward are essential elements of RL. 
In the context of AutoML, 
the environment is the generated machine learning solution, 
the reward corresponds to its performance on the holdout dataset, 
and the actions are operations to modify a learning solution. 

\begin{figure}
    \centering
     \includegraphics[width=0.33\textwidth]{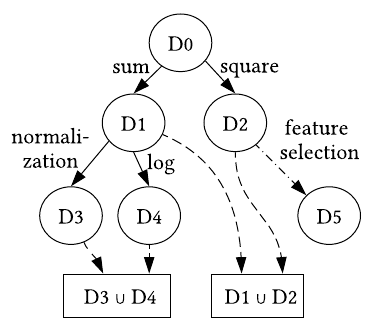}
    \caption{
    The transformation graph for AutoFE [18]. 
    $D_0$ is the initial dataset. The feature engineering operations such as sum, square, log, normalization, and feature selection, are used in the graph.
    }
    \label{fig:transformation_graph}
\end{figure}

Within the context of AutoFE, the environment corresponds to all the possible combinations of features and transformations, which could be represented in a transformation graph~\cite{khurana2018feature} (Figure~\ref{fig:transformation_graph}). 
The initial node of the graph is the original feature set. 
Edges are feature transformations (e.g., square, sum, and log), and feature selection. 
Other nodes are the derived features. 
Note that the transformation is applied to all features regardless of feature types. 
In the transformation graph, the traversal from the initial node to an end node is 
a path of feature transformation 
that applies all feature operators to the original dataset. 
The goal is to learn the traversal within the limited time budge to reach the optimal node brings about the highest performance.

The agent can learn the graph exploration with the help of Q-Learning, 
which uses Q function, as shown in Equation~\eqref{eq:q_learning} to remember previous learning experiences.
\begin{equation}
    \label{eq:q_learning}
        Q(s^{(t)}, a^{(t)}) \leftarrow \\
        (1 - \eta) Q(s^{(t)}, a^{(t)}) + 
        \eta[r^{(t)} + \gamma\max_{a^{'}}Q(s^{(t+1)}, a^{'})],
\end{equation}
where $s$, $a$, $\eta$, and $\gamma$ correspond to state, action, learning rate, and discount factor respectively and 
$r^{(t)}$ denotes the reward of step $t$. 
An action of the agent is a pair of an existing node $n$ in the current graph and a feature operation $t$ to the node.
The $\epsilon$-greedy algorithm determines the next action,
i.e., $(n,t)$ is chosen randomly with probability $\epsilon$ 
or from the maximum of Q function with probability $1-\epsilon$.
In each step, an action produces new features with which 
the learning algorithm $A$ is trained to get the validation performance.
The performance serves as the reward $r$ for the Q-function. 
Also, the current graph is augmented with $(n, t)$ and $r$.

Since the combination of states and feature operations grows exponentially in the graph, 
Q-Learning with linear approximation~\cite{khurana2018feature} is proposed.
That is, Q function approximates in $Q(s, a) = w^{a} f(s)$, 
where $w^a$ is a weight vector for an action $a$, and $f(s)$ is a state vector characterizing the state $s$. 
$w^{a}$ is updated by the following equation,
\begin{equation}
    \label{eq:q_approximation}
        w^{a_t} \leftarrow 
        w^{a_t} + 
        \alpha (r^{(t)} + \gamma\max_{a^{'}}
            [Q(s^{(t+1)}, a^{'}) - Q(s^{(t)}, a)]) f^{(t)}(s)
\end{equation}
where $\alpha$ is the learning rate. 
As long as the Q function is learned, the agent can decide the exploration direction.

\subsection{Evolutionary Algorithm}
\label{subsec:autofe_ea}
Evolutionary algorithm imitates the natural evolution to solve optimization problems, 
including genetic algorithms (GAs) and genetic programming (GP).
We will take AutoML using GP as an example in this survey, while other similar evolutionary algorithms such as particle swarm optimization (PSO) and ant colony optimization (ACO) will not be covered due to the limited space.

Typically, EA follows Algorithm~\ref{algo:ea}, including population, selection, crossover, and mutation. In the beginning, the evolution algorithms randomly generate individuals as the initial population. After evaluating the population, selection methods (e.g., tournament and elitism selection) will pick up competent individuals as parents to form the next population. In the mutation (crossover) step, mutation operations change parents' structure to create various individuals in the new population. The selection-mutation-crossover will be repeated until a particular condition is satisfied (e.g., a fixed number of generations).

Equipped AutoFE with EA, constructed features could be represented in the tree structure~\cite{tran2016genetic} where the root is a derived feature, the intermediate nodes are arithmetic operators, and leaves are either original features or random values. It is worth pointing out that the root of the tree is different from the transformation graph~\cite{khurana2018feature}, where the starting node is the original feature set. Genetic programming modifies the intermediate nodes and leaves of a tree-based individual to generate new features in the mutation and crossover phase. The selection phase of GP implicitly picks up candidate features. Therefore, the feature generation and selection are executed together within GP~\cite{tran2016genetic}.

\begin{algorithm}[]
    \label{algo:ea}
    \SetAlgoLined
    \SetKwInOut{Input}{Input}
    \SetKwInOut{Output}{Output}
    \Input{Individual representation; fitness measurement $\mathcal{M}$; Selection; Partition; Crossover; Mutation}
    \Output{The optimal individual}
     Initialize a population of random individuals, $P$ \\
     \While{not meet stopping criteria}{
        \For{individual $i$ in $P$}{
            $\mathcal{M}(i)$ \tcp*[l]{evaluate fitness of individual}
        }
        Generation $G$ = Selection($P$) \\
        $G_1$, $G_2$ = Partition($G$) \tcp*[l]{split $G$ to two sub-groups}
        $P$ = Crossover($G_1$) $\cup$ Mutation($G_2$) \\
     }
     \Return The individual with the highest fitness
     \caption{Evolutionary Algorithm}
\end{algorithm}

\subsection{Other Methods}
\label{subsec:autofe_other}
Besides RL and EA, 
there are also other approaches~\cite{kanter2015deep, 7837936, khurana2016cognito, 8215494} for AutoFE. The structure of datasets leads to the various techniques of feature generation. Meanwhile, the performance evaluation of selected features brings about varying techniques of feature selection. We review them in terms of feature generation and feature selection.

For feature generation, we can categorize datasets to relational and non-relational.
Relational datasets provide relationships between entities (tables) by following which we generate additional features~\cite{kanter2015deep}.
We apply arithmetic operators to columns of one entity 
that do or do not depend on columns of another entity. 
For instance, 
we can obtain a new feature of average customer expense by averaging the ``Order'' entity based on the IDs of the``Customer'' entity.
When it comes to non-relational datasets, we directly utilize the features in the dataset.
We can construct ridge and kernel ridge regression models to measure the feature correlations that serve as additional new features~\cite{8215494}. 
If datasets are expressed in the transformation graph (Figure~\ref{fig:transformation_graph}), depth-first and breath-first traversals~\cite{khurana2016cognito} without training an agent offer a simple way to decide the next node and feature transformation.

For feature selection, selected features could be evaluated by a tree model or a machine learning pipeline.
We can train a decision tree to get the information gain of each feature and use the stability-based selection that aggregates the results of many selection algorithms on different subsets of features. 
So mixed results are used to select final features~\cite{8215494}. 
Another way is constructing a random forest (RF) model over meta-features between datasets and new generated features~\cite{7837936}. 
The rank of generated features from the output of the RF model is used to select features. 
Apart from tree-based selections, 
a fixed-structure machine learning pipeline could be constructed for evaluation,
including Truncated SVD for feature selection~\cite{kanter2015deep}. 
We select $n_c$ components from the results of SVD transformation, rank the components by calculating their f-value w.r.t. the targets, and keep top $\gamma \%$ of them.
The \hypers $n_c$ and $\gamma$ are tuned by Bayesian optimization to select the number of features for the pipeline.

\begin{table*}[t!]
    \centering
    \tlstyle
    \label{table:comparison_auto_ml}
    \begin{threeparttable}
        \scalebox{0.9}{
            \begin{tabular}{|c|c|c|c|c|c|}  
            \hline
                \diagbox[width=10em]{Category}{Technique}
                 & \makecell{Bayesian \\ Optimization} & \makecell{Reinforcement \\ Learning} & \makecell{Evolutionary \\ Algorithm} & \makecell{Gradient-based \\ Approaches} & Other techniques \\
            \hline
            \makecell{Automated Feature\\ Engineering (AutoFE)} & 
                -
                & 
                \makecell{
                    FeatureRL~\cite{khurana2018feature} 
                } & 
                \makecell{GP for  Feature \\ Engineering~\cite{tran2016genetic}} &
                - & 
                \makecell{
                    Data Science Machine~\cite{kanter2015deep} \\
                    ExploreKit~\cite{7837936} \\ 
                    Cognito~\cite{khurana2016cognito} \\
                    AutoLearn~\cite{8215494}
                } \\
            \hline
            \makecell{Automated Model \\ and \Hyper \\ Learning (AutoMHL)} &
                \makecell{TPE~\cite{bergstra2011algorithms} \\ SMAC~\cite{hutter2011sequential}  \\ Auto-Sklearn~\cite{feurer2015efficient} \\ FABOLAS~\cite{klein2016fast} \\ BOHB~\cite{falkner2018bohb}} & 
                \makecell{
                    APRL~\cite{khurana2019automating} \\
                    Hyperband~\cite{li2016hyperband}
                } & 
                \makecell{TPOT~\cite{olson2016tpot} \\ Autostacker~\cite{chen2018autostacker} \\ DarwinML~\cite{qi2018darwinml}} & 
                \makecell{
                    -
                } & 
                -\\
            \hline
            \makecell{Automated Deep \\ Learning (AutoDL)} &
                \makecell{
                    AutoKeras~\cite{jin2018efficient} \\ 
                    NASBot~\cite{NIPS2018_7472} \\ 
                } & 
                \makecell{
                    NAS~\cite{zoph2016neural} \\ 
                    NASNet~\cite{Zoph2018CVPR} \\ 
                    ENAS~\cite{pmlr-v80-pham18a}
                } & 
                \makecell{
                    -
                } & 
                \makecell{
                    DARTS~\cite{liu2018darts} \\ 
                    Proxyless~\cite{cai2018proxylessnas} \\ 
                    NAONet~\cite{luo2018neural}
                } & 
                -\\
            \hline
            \end{tabular}
        }
    \end{threeparttable}
    \caption{Overview of different techniques for Automated Machine Learning.}
\end{table*}

\section{Automated Model and Hyperparameter Learning}
\label{sec: automhl}
%
Model and \hyper learning consists of model selection and \hyper tuning, which optimizes the predictive performance
by repeatedly changing machine learning models and tuning associate \hyper values. 
Given a dataset $\mathcal{D}$ divided into $\mathcal{D}_{train}$ and $\mathcal{D}_{val}$ for training and validation respectively, we should choose a loss function $\mathcal{L}_{train}$ for training (e.g., square mean error or cross entropy) and a measurement $\mathcal{L}_{val}$ for validation (e.g., accuracy or F1-score). Considering a set of learning models $\mathcal{A}=\{A^{(1)}, A^{(2)}, ...\}$, automated model and \hyper learning (AutoMHL) could be formulated as the following optimization problem,

\begin{equation}
    \label{eq:bilevel}
    \begin{aligned}
         A^{*}_{\lambda^*} & = \argmin_{A^{(i)} \in \mathcal{A}, \lambda \in \Lambda^{(i)}} \mathcal{L}_{val}(A^{(i)}_{\lambda, w^*}, D_{val}),
        \\
        \text{s.t.}\ \  w^* & = \argmin_{w} \mathcal{L}_{train}(A^{(i)}_{\lambda, w}, D_{train}),
    \end{aligned}
\end{equation}
where  $\Lambda^{(i)}$ denotes the \hyper space of $A^{(i)}$, 
$w$ is the parameters of a model $A$. The goal is to obtain the optimal learning model $A^{*}$ and its hyperparameters ${\lambda^*}$.
If we reduce $\mathcal{A}=\{A^{(1)}\}$, AutoMHL becomes \hyper optimization (HPO). If we expand $\mathcal{A}$ to include features $F_{i}$, i.e., $\mathcal{A}=\{A^{(1)}, A^{(2)}, ...F_1, F_2, ...\}$, AutoMHL becomes the automated pipeline learning (Auto-Pipeline). Therefore, AutoMHL is the general form of the above two AutoML problems. In this section, we briefly introduce BO and review AutoMHL in terms of BO, RL, and EA.

\subsection{Bayesian Optimization}
\label{subsec: automhl_bo} 

Bayesian optimization (BO) is a well-known method to find the optimal value of a black-box and non-convex objective function. 
It contains the two indispensable components, a probabilistic model and an acquisition function.
The probabilistic model is a surrogate model emulating the objective function $f$, which models the functional relationship between the search space and the objective function values.
The acquisition function is constructed based on the surrogate to measure potential points in the search space by considering the trade-off between exploration (trying uncertain points) and exploitation (improving current good points). 
BO goes through an iterative procedure of three steps, as shown in Algorithm~\ref{algo:bo}.
(1) Line~\ref{alg:opt_alpha}: obtain the next point to evaluate by optimizing the acquisition function. (2) Line~\ref{alg:eval}: evaluate the point to get its function value. (3) Line~\ref{alg:augment}: update the surrogate model with the new point and the value. 



There are two challenges to be considered when applying BO for AutoMHL, including heterogeneous search space and efficiency.
The first challenge results from various \hypers of a learning model, including
discrete, categorical, continuous hyperparameters, and hierarchical dependency (conditional) among them. 
Take support vector machines (SVM) as an example.
Its kernel, degree, and gamma is a categorical, discrete, and continuous \hyper respectively, while the degree also depends on polynomial kernel. 
The efficiency bottleneck usually arises from the performance evaluation step. In every iteration, we need to obtain the real value of the objective function. The situation becomes worse when it is costly to train a learning model (e.g., neural network) on a large-scale dataset. Therefore, various techniques have been proposed to address these challenges.

First, tree-based BO~\cite{hutter2011sequential, bergstra2011algorithms} have emerged as alternates to handle the heterogeneous search space.
Gaussian process in standard BO cannot consider the hierarchical dependency 
and model the probabilities of various types of hyperparameters.
Random forest (RF) regression could replace it as the surrogate model~\cite{hutter2011sequential}.
Random subsets of historical observations build individual decision trees.
One tree node decides the value to split a discrete/continuous \hyper or 
whether a \hyper is active to handle conditional variables. 
To build an acquisition function (e.g., Expected improvement) upon the RF regression, 
we have to measure the probability $p(c|\lambda)$ of real performance $c$
given \hypers $\lambda$.
The RF regression does not produce $p(c|\lambda)$, but
the frequency estimates of $\mu_\lambda$ and $\sigma^2_\lambda$ from individual trees could be used to 
approximate $p(c|\lambda) = \mathcal{N}(\mu_\lambda, \sigma^2_\lambda)$.

Another way to replace Gaussian process is by building a probability density tree~\cite{bergstra2011algorithms},
in which a tree node stands for the probability density of a specific \hyper rather than a decision.
A \hyper serves as a random variable.
There are two types of probability density in a node, 
\begin{equation}
    \label{eq:tpe}
    \begin{aligned}
        p(\lambda | c ) = \begin{cases}
            l(\lambda) = p(\lambda | c < c^*) \\
            g(\lambda) = p(\lambda | c >= c^*).
        \end{cases}
    \end{aligned}
\end{equation}
The probability density of \hypers $\lambda$ conditional on 
whether the performance $c$ is less or greater than a given performance threshold $c^*$.
For the dependent hyperparameters, 
only values of active ones update the probability of the corresponding nodes.
Since the density tree uses $p(\lambda | c )$ instead of $p(c|\lambda )$, Bergstra et al.~\cite{bergstra2011algorithms} show that the optimization of Expected improvement could be simplified by
$\lambda^* = \argmin_\lambda\frac{g(\lambda)}{l(\lambda)}$. 

In the context of Auto-Pipeline, we can expand the search space~\cite{feurer2015efficient} by adding pipeline operations, 
such as data imputation, feature scaling, transformation, and selection.
The tree-based BO~\cite{hutter2011sequential, bergstra2011algorithms} could find the optimal pipeline solution in the new search space.
However, the new space is much larger than that of HPO.
Considering past performance on similar datasets~\cite{feurer2015efficient} 
enables BO to start in the small but reasonable regions, 
which reduces the number of intermediate results. 
For those sub-optimal pipelines during a search, 
we can also build a weighted ensemble of them~\cite{feurer2015efficient} to get robust performance.

\begin{algorithm}[]
    \label{algo:bo}
    \SetAlgoLined
    \SetKwInOut{Input}{Input}
    \SetKwInOut{Output}{Output}
    \Input{Objective function $f$; surrogate model $\mathcal{M}$; acquisition function $\alpha$; current observations \ $\mathcal{D}_n = \{(\mathbf{x}_i, y_i)\}_{i=1}^{n}$};
    \Output{The optimal value of $f$}
     Initialize $\mathcal{M}$ and $\alpha$ \\
     \While{not meet  criteria}{
        $\mathbf{x}_{n+1} = \argmax_{\mathbf{x}} \alpha(\mathbf{x};  \mathcal{D}_n)$ \\ \label{alg:opt_alpha}
        $y_{n+1} = f(\mathbf{x}_{n+1})$ \\ \label{alg:eval}
        $\mathcal{D}_{n} = \mathcal{D}_n \cup \{(\mathbf{x}_{n+1}, y_{n+1} )\}$ \\
        update $\mathcal{M}$ with $\mathcal{D}_{n}$ \\ \label{alg:augment}
     }
     \Return The optimal value of $f$ according to $\mathcal{D}_{n}$
     \caption{Bayesian optimization}
\end{algorithm}

Second, 
approximate performance evaluation (multi-fidelity optimization) is proposed to decrease the evaluation cost.
Specifically, we could train the model learning in an economical setting.
For example, we can use a subset of training data~\cite{nickson2014automated} or fewer epochs of stochastic gradient descent~\cite{swersky2014freeze} when training neural networks. 
Performance extrapolation is an alternative acceleration method by extrapolating the performance of the full dataset based on the historical records of the partial datasets. 
For instance, we can prepare two objective functions, 
the loss function for performance evaluation and the training time function, both of which consider the size of the training set as an additional input~\cite{klein2016fast}.
In individual iterations, Bayesian optimizer determines the proper size of training data that could take a few training time and ideally represent the performance of the full training set as well. 
Thus, the approach could allow training the learning model with a small dataset in early stages and extrapolating the performance of the full set.

Dynamic resources assignment is an approach assigning more training budgets for promising configurations and fewer budgets for discouraging configurations. Hyperband~\cite{li2016hyperband} is a representative example illustrated in Section~\ref{subsec:automhl_rl}.
An incremental work BOHB~\cite{falkner2018bohb} using Hyperband to speed up a tree-based BO approach~\cite{bergstra2011algorithms}. 
Different from Hyperband~\cite{li2016hyperband} sampling configurations by random search, BOHB relies on a model-based approach (TPE)~\cite{bergstra2011algorithms} to select encouraging configurations, making it easier to converge the performance of resulting configurations.

\begin{figure}
    \centering
     \includegraphics[width=0.42\textwidth]{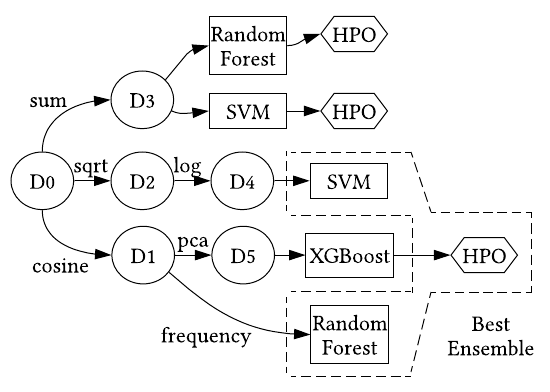}
    \caption{
        The exploration tree for AutoMHL with RL [17].
        The root $D_0$ is the initial dataset, and circular nodes are derived datasets by applying a feature transformation. 
        Rectangular nodes stand for machine learning algorithms. 
        HPO indicates whether we tune \hypers of the learning model. 
        The goal is to find the ensemble of learning models for a machine learning pipeline.
    }
    \label{fig:exploration_tree}
\end{figure}

\subsection{Reinforcement Learning}
\label{subsec:automhl_rl}


Speaking of AutoMHL with RL, the agent is required to determine the \hypers of a learning algorithm (HPO)~\cite{li2016hyperband} 
or design a machine learning pipeline (Auto-Pipeline)~\cite{khurana2019automating}. 
We review the RL techniques from the aspect of HPO and Auto-Pipeline.


For HPO, we could frame it in the multi-armed bandit, a particular case of RL.
The environment consists of multiple arms, and 
the agent pulls a sequence of the arms to obtain the maximal reward. 
A set of \hypers stands for an arm
whose reward is the performance of a learning model with the \hypers\cite{li2016hyperband}. 
The goal is to find the optimal arm by pulling a sequence of arms one-by-one. 
Given $n$ arms ($n$ sets of hyperparameters), we can quickly determine the optimal one
as long as getting their performance.
To efficiently obtain the performance under a fixed budgets $B$, 
successive halving~\cite{jamieson2016non} assigns each configuration $\frac{B}{n}$ budgets
and then select the best half with double budgets in the next iteration
until one configuration remains.
Furthermore, 
we could dynamically adjust budgets (e.g., the number of iterations) according to $n$~\cite{li2016hyperband}; 
large $n$ configurations occupy few budgets while few $n$ configurations consume large budgets. 
The two-fold dynamic resource allocation significantly reduces the evaluation time~\cite{li2016hyperband}. 
Thus, when $n$ arms are sampled by random search~\cite{li2016hyperband} or tree-based BO~\cite{falkner2018bohb}, the aforementioned multi-armed techniques could efficiently determine the best one.

For Auto-Pipeline, 
we could picture its search space to the exploration tree~\cite{khurana2019automating} (Figure~\ref{fig:exploration_tree}), 
where the root is the original dataset with raw features, 
intermediate nodes are derived features after feature transformation, and 
the leaves are HPO or learning models.
HPO indicates whether to tune \hypers of the parent node.
The agent selects an existing node and applies a transformation, a learning model, or HPO to the node.
The goal of RL is to learn an exploration policy in the tree
so that selected learning models could form an ensemble providing the best predictive performance.
The agent takes advantage of Q-Learning to remember historical rewards and adopts $\epsilon$-greedy algorithm to decide the next node and the operation.
To avoid the exponential combinations of nodes and operations, Q-function is approximated by 
$Q(s, c) = w*f(s)$~\cite{khurana2019automating},
where $s$, $c$, and $w$ denote a state, an action, and a weight vector respectively.
$f(s)$ is a state vector characterizing the state $s$. 
Following the Equation~\eqref{eq:q_approximation}, we can learn the Q-function to decide actions in the tree. 

\begin{figure}
    \centering
    \includegraphics[width=0.5\textwidth]{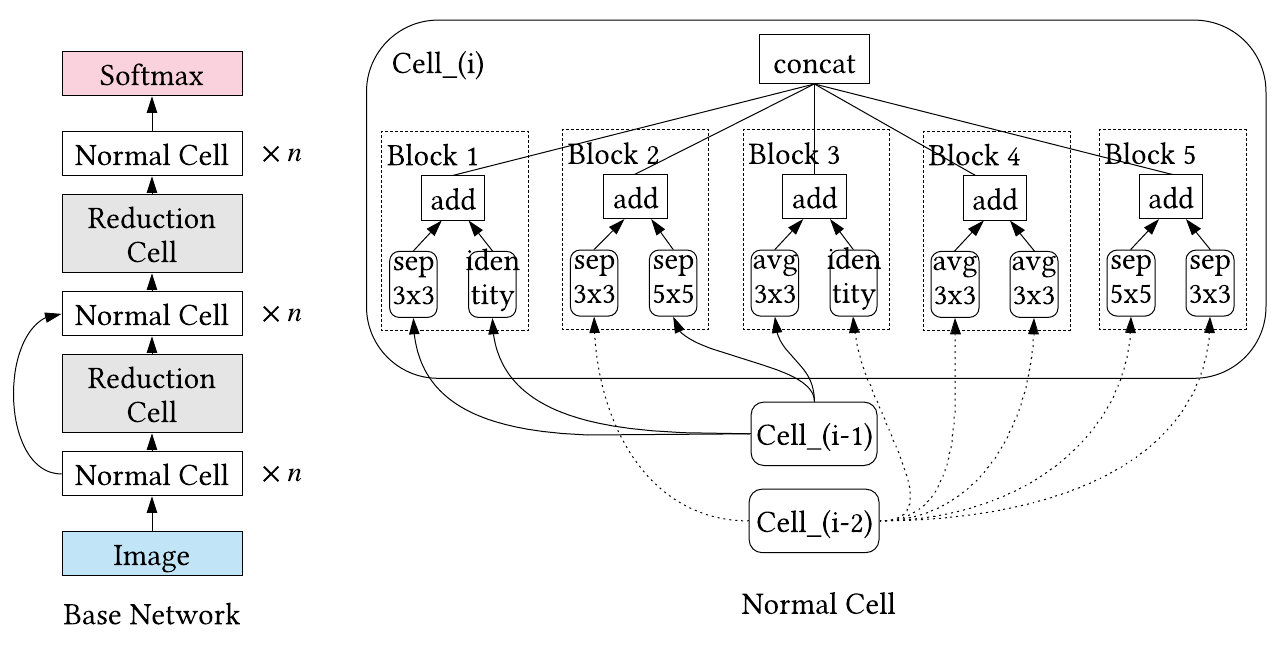}
    \caption{
    An example of a cell-based search space [35]. 
    The left figure is a base network that stacks normal cells and reduction cells with skip connections.
    The right one is a normal cell, a convolutional cell 
    that returns a feature map of the same dimension.
    A reduction cell is also a convolutional cell with a large stride size,
    which might have the same architecture as the normal cell.
    Both types of cell are composed of multiple blocks where each block has two operations and two inputs, 
    and these operation results are added.
    The sep, avg, and identity denote 
    depthwise-separable convolution, average pooling, and $1\times1$ convolution, respectively.
    In the search space, we are supposed to determine the two cell architectures and construct the base network to evaluate the performance of resulting cells.
    }
    \label{fig:cell_search}
\end{figure}

\subsection{Evolutionary Algorithm}
\label{subsec:automhl_ea}
Extant AutoMHL work with EA~\cite{olson2016tpot, chen2018autostacker, qi2018darwinml} focus on automated pipeline learning (Auto-Pipeline). 
In particular, they utilize genetic programming (GP) to search machine learning pipelines. 
GP encodes pipelines in flexible graph structures and evolves them into the optimal one that provides the highest predictive performance. 
There are three types of flexible pipeline structures, 
tree-based~\cite{olson2016tpot}, layer-based~\cite{chen2018autostacker}, and graph-based pipeline~\cite{qi2018darwinml}. 

First, in a tree-based structure~\cite{olson2016tpot}, leaves are the copies of input data,
intermediate nodes are pipeline operators, such as feature pre-processing methods, feature selection algorithms, and machine learning models and
the root node  must be a learning model.
A tree pipeline has one or more pre-processing nodes followed by a selection node and a model node.
The mutation and crossover of GP might change the types of pipeline operators as well as \hypers of each pipeline operators.
The fitness is the predictive performance of a model node after intermediate nodes process input data.

Second, a layer-based pipeline~\cite{chen2018autostacker} is similar to multiple layer perceptron, 
but each node is a machine learning model.
The first layer is the input data, 
and the outputs of the following layer are new synthetic features added to the raw features that serve as the input of the next layer.
The search space contains the number of layers, the number of nodes per layer, types of learning models, and associate \hyper of each learning model. 
GP tunes the above space to find the optimal pipeline.

Finally, the most general pipeline structure is a directed acyclic graph (DAG)~\cite{qi2018darwinml},
where edges and vertices indicate data flow and learning model respectively.
A pipeline is able to combine many various types of machine learning models, such as the mix of classifiers, regressors, and unsupervised learning models.
The adjacent matrix controls the edge connections of a DAG, which is evolved by mutation operators. After GP designs the pipeline structure and learning models, associate \hypers could be tuned by Bayesian optimization.
The last two pipeline structures do not explicitly operate feature engineering, but they leverage outputs of learning models as new features.

\section{Automated Deep Learning}
\label{sec:autodl}

Automated deep learning (AutoDL) is purposed to facilitate the design of neural architectures and the selection of their hyperparameters. 
It can be regarded as a particular topic of AutoMHL. Following Equation~\eqref{eq:bilevel}, we can describe AutoDL in the following optimization problem,
\begin{equation}
    \label{eq:autodl}
    \begin{aligned}
         A^{*} & = \argmin_{A \in \mathcal{A}} \mathcal{L}_{val}(A(w^*), D_{val}),
        \\
        \text{s.t.}\ \  w^* & = \argmin_{w} \mathcal{L}_{train}(A(w), D_{train}),
    \end{aligned}
\end{equation}
where $A$ is a network architecture, and $\mathcal{A}$ is the search space. 
The goal is to search an optimal network architecture $A^{*}$ that could achieve the highest generalization performance on the validation set. 

Generally, search space $\mathcal{A}$ includes the following choices, (1) layer operations, e.g., perceptron, convolution, and max pooling, (2) layer hyperparameter, e.g., the number of hidden units, filters, and stride size, and (3) skip connections. 
We use these choices to build either the entire network architectures or cell architectures, resulting in whole-network search space and cell-based search space~\cite{Zoph2018CVPR}. The former is used to construct the entire network architecture from scratch. The latter aims to design the repeated cells (normal cells and reduction cells)
as shown in Figure~\ref{fig:cell_search} and construct the entire network architecture by stacking these cells. Due to their popularity and generality, in the rest of this section, we illustrate mainstream searching techniques that are proposed upon them.


\subsection{Bayesian Optimization}
\label{subsec:autodl_bo}
When it comes to Bayesian optimization (BO) for AutoDL, there are two challenges required to overcome: (1) how to measure the similarity of two neural architectures, which are not defined in Euclidean space, and (2) how to optimize the acquisition function. 
The former arises from the standard BO that relies on Gaussian process as the surrogate of the objective function.
Gaussian process requires measurable distance between two points to introduce a kernel function. The latter challenge appears due to discrete architectural choices in the search space,
which makes it hard to optimize the acquisition function to get candidate architectures.

To resolve the first challenge, 
we could design a similarity measurement for two arbitrary architectures in the search space~\cite{jin2018efficient, NIPS2018_7472}. 
Network edit-distance~\cite{jin2018efficient} is proposed to quantify the similarity of two networks. 
It calculates the number of network morphism operations~\cite{chen2015net2net} required to transform one network to another 
and introduces a valid kernel function based on the Bourgain embedding algorithm. 
Another similarity measurement, layer matched mass~\cite{NIPS2018_7472}, 
measures the amount of matched computation at the layers of one network to the layers of the other in terms of the matched frequency and the location of layer operations. 
The corresponding distance is defined as the minimum of layer matched mass computed by an Optimal Transport (OT) program, that finds the optimal transportation plan via solving a cost minimization problem.





To address the second challenge, recent work~\cite{jin2018efficient} proposes $A^*$ search with simulated annealing to optimize the acquisition function in the tree-structured search space. 
As network morphism~\cite{chen2015net2net} enables the transformation among different networks, 
the search space could be defined as a tree-structured graph, in which
nodes are neural architectures, 
and edges denote morphism operations changing the architecture of the parent node.
$A^*$ search conducts exploitation
by expanding the architecture of the best node in the tree 
while simulated annealing performs exploration, which randomly chooses other nodes for expansion with certain predefined probabilities. Evolutionary algorithm~\cite{NIPS2018_7472} is also proposed to address the challenge of searching in discrete space. It mutates $N$ neural architectures, which have higher acquisition values by random $N$ layer operations,
evaluates the acquisition of the new ones, 
and repeats the above steps until the prescribed condition.
Upon solving the two challenges, we could follow the standard BO procedure to solve Equation~\eqref{eq:autodl}.

\begin{figure}
    \centering
    \includegraphics[width=0.4\textwidth]{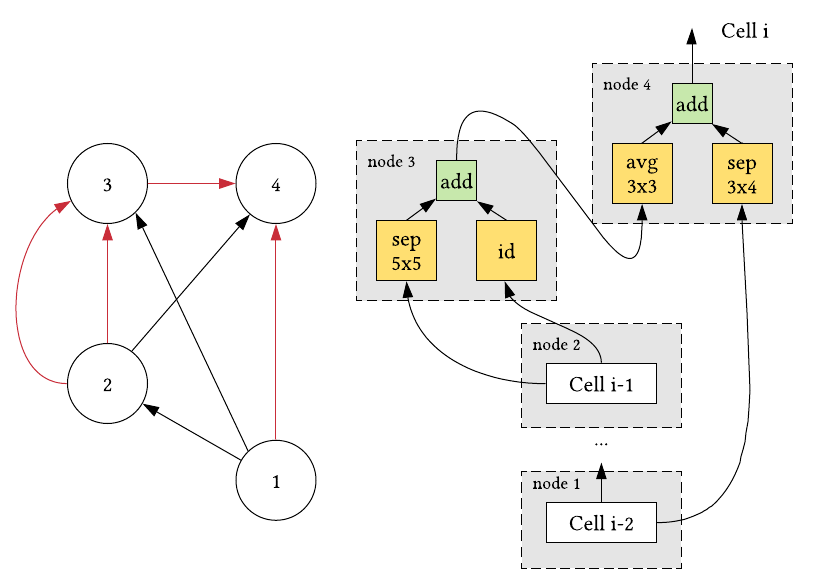}
    \caption{
    An example of a directed acyclic graph (DAG) for AutoDL with weight sharing [26]. 
    The DAG is built on cell-based search space and has four nodes, 
    where node 1 and node 2 are two previous cells' outputs, 
    and node 3 and node 4 are blocks of the current cell.
    The four nodes and active edges (red) make up a sub-graph for a cell architecture.
    All sub-graphs share weights of layer operations used in cell neural networks.
    }
    \label{fig:enas_dag}
\end{figure}

\subsection{Reinforcement Learning}
\label{subsec:autodl_rl}

Within the context of AutoDL with reinforcement learning (RL), the agent is required to learn a policy to construct network architectures. 
The trained networks will produce validation accuracy as reward $R$ for the agent.
We explain how to train the agent in terms of two kinds of search space. 

For the whole-network search space, the agent generates architectural \hypers of the entire neural networks.
One way to train the agent is policy gradient~\cite{zoph2016neural}. 
The agent is modeled by Recurrent Neural Network (RNN) which predicts a variable string to depict the architectural hyperparameters.
The policy of the RNN agent is formulated as 
the probability $P(a_{1:T};\theta_c)$ of a sequence of $T$ actions for $T$ \hypers
given the RNN parameters $\theta_c$. 
It could be iteratively updated via policy gradient towards maximizing the expected reward $J(\theta_c) = E_{P(a_{1:T};\theta_c)}[R]$.
The method~\cite{zoph2016neural} following REINFORCE rule expresses the reward gradient $\nabla_{\theta_c}J(\theta_c)$ as follows,
\begin{align}
\label{eq:approx_pg}
    \nabla_{\theta_c}J(\theta_c) = \frac{1}{m}\Sigma_{k=1}^{m}\Sigma_{t=1}^{T}
    \nabla_{\theta_c}\ln{P(a_t|a_{(t-1):1};\theta_c)}
    R_k,
\end{align} 
where $m$ is the number of sampled architectures in one batch, and $R_k$ means the validation accuracy of neural network $k$. 
After getting the validation accuracy of a batch of architectures, we update the agent's policy by Equation~\eqref{eq:approx_pg}.

For the cell-based search space, 
the agent only searches the architecture of normal and reduction cells.
The revised RNN-agent~\cite{Zoph2018CVPR} predicts a string of inputs and operations of each block within a cell.
By transforming the cell descriptions into architectures
and stacking them according to the base network, a complete network architecture is obtained and evaluated to provide rewards for the agent. 
For the agent training,
another policy gradient method, Proximal Policy Optimization (PPO)~\cite{Zoph2018CVPR}, could be adopted.

Moreover, 
weight sharing~\cite{pmlr-v80-pham18a} is an acceleration technique to search neural architectures with RL in two kinds of search space. 
It constructs a large directed acyclic graph (DAG)
where all sub-graphs can represent all neural networks in a particular search space.
Hence, it forces all child models to share parameters (weights)
and avoids training each model from scratch to converge.
For example, 
Figure~\ref{fig:enas_dag} presents the DAG of a cell-based search space. 
Each node corresponds to a block of a cell,
and directed edges represent how blocks are connected.
A sub-graph is a cell architecture.
The RNN agent is trained by policy gradient to return a string of inputs and operations for each block,
that determines node computations and connected edges in the large graph. 
When we train a complete network using the cell description, 
weights of a layer operation corresponding to the node in the large graph are updated.
Therefore, weights are shared with several child models.

\begin{table*}
    \centering
    \label{table:software}
    \begin{threeparttable}
        \scalebox{0.92}{
            \begin{tabular}{|c|c|c|c|c|}  
                \hline
                Group & Framework & Language & Provided by & URL\\
            \hline
            \multirow{2}{*}{\makecell{Automated Feature\\ Engineering (AutoFE)}} 
                & FeatureTools~\cite{kanter2015deep} & Python & Open-Source & https://github.com/Featuretools/featuretools \\ 
                \cline{2-5}
                & AutoCross~\cite{yuanfei2019autocross} & - & 4paradigm & https://www.4paradigm.com \\
            \hline
            \multirow{8}{*}{\makecell{Automated Model \\ and \Hyper \\ Learning (AutoMHL)} }
                & Hyperopt~\cite{bergstra2011algorithms} & Python & Open-Source & https://github.com/hyperopt/hyperopt\\ 
                \cline{2-5}
                & SMAC3~\cite{hutter2011sequential} & Python & Open-Source & https://github.com/automl/SMAC3 \\ 
                \cline{2-5}
                & Auto-Sklearn~\cite{feurer2015efficient} & Python & Open-Source & https://github.com/automl/auto-sklearn \\ 
                \cline{2-5}
                & TPOT~\cite{olson2016tpot} & Python & Open-Source & https://github.com/EpistasisLab/tpot\\
                \cline{2-5}
                & HyperTune & - & Google & https://bit.ly/2IMsECx \\
                \cline{2-5}
                & Automatic Model Tuning & - & Amazon & https://aws.amazon.com/sagemaker \\
                \cline{2-5}
                & Azure AutoML  & - & Microsoft & https://bit.ly/2XxBMmA \\
                \cline{2-5}
                & DarwinML & - & Intelligence Qubic & http://iqubic.net/ \\
            \hline
            \multirow{3}{*}{\makecell{Automated Deep \\ Learning (AutoDL)}}
                & Auto-Keras~\cite{jin2018efficient} & Python & Open-Source &  https://github.com/keras-team/autokeras \\ 
                \cline{2-5}
                & AdaNet~\cite{weill2019adanet} & Python & Open-Source & https://github.com/tensorflow/adanet \\
                \cline{2-5}
                & Google AutoML & - & Google & https://cloud.google.com/automl \\
                \cline{2-5}
                & Neural Network Intelligence & Python & Microsoft & https://github.com/microsoft/nni \\
            \hline
            \end{tabular}
        }
        \caption{A list of selective AutoML Frameworks until June 2019}
    \end{threeparttable}
\end{table*}

\subsection{Gradient-based Approach}
For gradient-based approaches with AutoDL, 
the most critical challenge is to convert non-differentiable optimization problems into differentiable ones. 
Thus, we want to encode a network architecture into a continuous and numeric representation. 
Once having continuous representation, we can optimize the architectural \hypers with gradient descent, which is the same procedure to optimize the parameters of neural networks.
Generally, there are two ways for the transformation.

One way is the mixed operation encoding~\cite{liu2018darts, cai2018proxylessnas}. Given a candidate set of layer operations $\mathcal{O} = \{o_{i}\}$ and an input $x$, the output of a mixed operation $m_{\mathcal{O}}(x)$~\cite{liu2018darts} is a weighted sum of these operations.
\begin{align}
\label{eq:mixed_operation}
    m_{\mathcal{O}}(x) = \sum_{o \in \mathcal{O}} 
        \frac{\exp{(\alpha_{o})}}{\sum_{o^{i} \in \mathcal{O}}\exp{(\alpha_{o^{i}})}} o(x),
\end{align} 
where $\alpha_{o}$ is the numerical weight for the operation $o$, called architectural weight.
In the cell-based search space, 
each edge of a cell network is specified a mixed operation, and 
a cell with $E$ edges is then encoded in $|\mathcal{O}| \times E$ architectural weights.
Furthermore, the mixed operation can be improved its memory usage by binary gates~\cite{cai2018proxylessnas}, 
where an operation is sampled according to the probability proportional to its architectural weight. 
Thus, only one operation in an edge will be activated during learning rather than computing the weights of all operations. 
The goal of the representation is using gradient approaches to learn these architectural weights, 
like running 1st-order gradient descent alternatively for Equation~\eqref{eq:autodl}.
It fixes the outer architectural weights to update the gradient of inner weights
and then fixes the inner weights to update the gradients of outer architectural weights. 
After learning, the largest value of architectural weights for an edge decides its operation.  

Another feasible way is using encoder/decoder to learn continuous architecture representation~\cite{luo2018neural}. 
The encoder transforms an architectural string to an embedding representation,
and the decoder reverts an embedding to a string. 
This representation combines a performance predictor.
It maps an embedding to its validation performance in the latent space 
where stochastic gradient descent attempts to search the embedding representation with the best performance.

\section{AutoML Frameworks}
\label{sec:frameworks}
In this section, we discuss representative AutoML frameworks from either open-source project or enterprise services. Note that we exclude those open-source codes that are merely experiment implementations for research papers. 

\subsection{Automated Feature Engineering}
FeatureTools~\cite{kanter2015deep} is an open-source framework using Python which can automatically generate features from relational datasets. The generation follows the relationship of any two entities to apply feature transformation operators, such as average and count. 
Besides, AutoCross~\cite{yuanfei2019autocross} is a feature crossing tool provided by 4Paradigm. The tool automatically captures interactions between categorical features by cross-product. Its cross feature generation relies on beam search to generate candidate features. Feature selection is ranking the performance of features on a logistic regression regardless of the real learning model. 


\subsection{Automated Model and Hyperparameter \\ Learning}
Current AutoMHL frameworks are split into HPO and Auto-Pipeline developed from the open-source community or enterprise.
We introduce them in the light of open-source and enterprise.

For open-source frameworks, 
Hyperopt~\cite{bergstra2011algorithms} and SMAC3~\cite{hutter2011sequential} are tools for HPO, 
while Auto-Sklearn~\cite{feurer2015efficient}, TPOT~\cite{olson2016tpot}, and H2O are frameworks for Auto-Pipeline.
Hyperopt~\cite{bergstra2011algorithms} and SMAC3~\cite{hutter2011sequential} use Bayesian optimization to tune \hypers of a learning model.
Auto-Sklearn~\cite{feurer2015efficient} is the extension of SMAC~\cite{hutter2011sequential} with meta learning and ensemble learning for machine learning pipeline automation. 
TPOT~\cite{olson2016tpot} is the most popular Auto-Pipeline tool using genetic programming to optimize the tree-structure pipeline. 
H2O is an open-source machine learning platform written in Java developed by H2O.ai.
The platform contains solutions of pipeline automation by optimizing \hypers via random search, grid search, and Bayesian optimization and constructing stacking ensembles.

For enterprise frameworks, 
Google and Amazon provide a cloud service for HPO, 
while Microsoft and Intelligence Qubic offer Auto-Pipeline frameworks.
Google HyperTune could tune \hypers of machine learning models by grid and random search, TPE~\cite{bergstra2011algorithms},
and SMAC~\cite{hutter2011sequential}.
Amazon Automatic Model Tuning could optimize \hypers using Bayesian optimization.
Furthermore, Microsoft Azure AutoML uses collaborative filtering and Bayesian optimization to search for promising pipelines. 
DarwinML is an Auto-Pipeline platform provided by Intelligence Qubic.
It relies on genetic programming to construct arbitrary architectures of ensemble models for a machine learning pipeline.

\subsection{Automated Deep Learning}
AutoDL is a subfield of AutoMHL, but AutoDL frameworks mainly build the architectures of neural networks automatically.
We introduce two well-known open-source frameworks, Auto-Keras and AdaNet,
and two enterprise services,  Google AutoML and Neural Network Intelligence (NNI) from Microsoft.

Auto-Keras~\cite{jin2018efficient} is an open-source project using Keras to search for deep network architectures. 
Its adopts Bayesian optimization and network morphism to search entire arbitrary neural architectures instead of cell network architectures. Moreover, AdaNet~\cite{weill2019adanet} is another open-source work using Tensorflow for learning network architecture automatically. It could learn network architectures and build an ensemble of networks to obtain a better model. 

Besides,
Neural Network Intelligence (NNI) is an open-source toolkit provided by Microsoft for neural architecture search and \hyper tuning. 
It supports a various of search techniques, e.g., evolutionary algorithm and network morphism,
to tune network architectures
and provides flexible execution environments in local or cloud servers.
Last but not least, 
Google Cloud AutoML is a service to get high-quality neural networks on vision classification, text classification, and language translation. 
The service is based on the NAS proposed by Zoph and Le~\cite{zoph2016neural}. 
Users could obtain customized models by merely uploading their labeled datasets without any pieces of code.

\section{Challenges}
\label{sec:challenge}
Authoritative benchmarks are essential standard protocols for AutoML comparison
that regulate datasets, the search space, and the training setting of searched machine learning solutions.
Unfortunately, such benchmarks are not omnipresent in AutoML.
No standard protocol for AutoFE provides a fixed set of transformation operations, and identical training setting of generated solutions.
HPOLib~\cite{eggensperger2013towards} collects a set of benchmarks for HPO, but no benchmark exists for Auto-Pipeline.
Researchers usually leverage the UCI Machine Learning Repository and OpenML datasets with their setting for evaluation.

For AutoDL, NAS-Bench-101~\cite{ying2019bench} is a tabular benchmark 
which maps five millions of trained neural architectures to their training and evaluation metrics, 
where these architectures are generated from the cell-based search space. 
As the objective comparison, 
any AutoDL technique generates child networks within the same search space 
and looks up their performance without training them from scratch,
which avoids different training techniques and hyperparameters.
However, the NAS-Bench-101~\cite{ying2019bench} only considers the cell-based search space. 
For the comparison of AutoDL within whole-network search space, 
one of the first attempts is Auto-Keras~\cite{jin2018efficient}, 
which encompasses data preparation, training setting of child networks, and optimizer.
Researchers can concentrate on inventing the AutoDL optimizer, 
and Auto-Keras~\cite{jin2018efficient} will produce the empirical performance in the standard testing environment. 
Authoritative benchmarks for AutoML are required; 
otherwise, it would be hard to compare the effectiveness and efficiency among a multitude of AutoML techniques.

Efficiency is one of the dominant factors for the successful adoption of AutoML methods. Some AutoML methods only take tolerable time to generate solutions: Auto-Sklearn~\cite{feurer2015efficient} requires 500 seconds in a CPU to find competence pipelines; FeatureRL~\cite{khurana2018feature} takes 400 seconds in a CPU to find appropriate features for a learning model. In contrast, AutoDL usually demands massive computation resources and time, which could consume up to thousands of GPU days~\cite{zoph2016neural}. Although novel acceleration techniques are proposed to reduce the enormous order of time~\cite{jin2018efficient, pmlr-v80-pham18a}, the efficiency of NAS is still an open problem.

The design of the search space is another challenge for AutoML, 
which requires substantial knowledge and experience from human experts.
A practical search space incorporates human prior knowledge for machine learning 
and considers the trade-off between its size (compact) and all useful possibilities (comprehensive).
A well-known example is the cell-based search~\cite{Zoph2018CVPR} for AutoDL.
The repeated neural network blocks in ResNet or DenseNet motivate
researchers to fix the base network architecture and to search the cell typologies only.
The handcrafted search spaces highly affect the performance of AutoML, which is still an intriguing question. 


Interpretable analysis can provide insight on what crucial components from AutoML could lead to desired performance. With the integration of friendly user interfaces, AutoML users could also customize their solutions by selecting desirable configurations between independent runs. 
Nevertheless, any AutoML search techniques is still a black-box procedure. 
Thus, limited search knowledge from AutoML can be utilized interactively. We believe interpretability and user interaction can increase the usefulness and capability of AutoML.


Miscellaneous domains remain potential to collaborate with AutoML. Most AutoML work have focused on classical regression and classification problems, particularly AutoDL whose empirical results are majorly from image classification on CIFAR-10. Nevertheless, the optimization setup of other potential domains more or less differs from the conventional AutoML. Take anomaly detection as an example. The extremely skew distribution of the benign class against the anomaly class might lead AutoML to search biased machine learning solutions. Other domains must have distinct considerations. We believe the cooperation of AutoML with multiple potential application domains could become another promising research direction.

\section{Conclusion}
\label{sec:conclusion}

Automated machine learning (AutoML) is the end-to-end process, which automatically discovers machine learning solutions. 
Without laborious human involvement, 
AutoML can expedite the process of applying machine learning to specific domain problems. 
Apart from our work,
there are also other AutoML reviews~\cite{elsken2019neural, quanming2018taking}, unveiling the mysterious art behind AutoML. 
As most of us develop with respect to three indispensable elements: 
search space, search techniques, and performance evaluation techniques, what makes us distinctive is that we provide a new perspective via categorizing AutoML into AutoFE, AutoMHL, and AutoDL, and elucidating them from technique dimension and framework dimension. 
By including and introducing both mainstream AutoML techniques and selected AutoML frameworks,
we hope this survey could give insight into the progress of AutoML. 



\bibliographystyle{abbrv} 
\bibliography{bibs4cacm}  
%
\balancecolumns
\end{document}